\documentclass[runningheads]{llncs}

\usepackage[T1]{fontenc}
\usepackage{makecell}
\usepackage{graphicx,verbatim, amsmath, amsfonts}
\usepackage{printlen}
\usepackage{booktabs}
\usepackage[justification=centering]{caption}
\usepackage{xcolor}
\usepackage[
    colorlinks=true,
    linkcolor=blue,      % color of internal links (e.g., \ref, \toc)
    citecolor=blue,      % color of citations (\cite)
    urlcolor=blue        % color of external URLs
]{hyperref}

\begin{document}

\title{Implicit Neural Representations of Intramyocardial Motion and Strain}

\author{Andrew Bell\inst{1} \and
Yan Kit Choi\inst{1} \and
Steffen E Petersen\inst{2} \and Andrew King\inst{1} \and Muhummad Sohaib Nazir\inst{1,3} \and Alistair A Young\inst{1}}
\authorrunning{A. Bell et al.}
\institute{School of Biomedical Engineering and Imaging Sciences, King's College London, United~Kingdom\and
William Harvey Research Institute, Queen Mary University of London, United~Kingdom \and Cardio-Oncology Service, Royal Brompton and Harefield Hospitals, London, United~Kingdom}

\maketitle    
\begin{abstract}
Automatic quantification of intramyocardial motion and strain from tagging MRI remains an important but challenging task. We propose a method using implicit neural representations (INRs), conditioned on learned latent codes, to predict continuous left ventricular (LV) displacement --- without requiring inference-time optimisation. Evaluated on 452 UK Biobank test cases, our method achieved the best tracking accuracy (2.14 mm RMSE) and the lowest combined error in global circumferential (2.86\%) and radial (6.42\%) strain compared to three deep learning baselines. In addition, our method is  $\sim380\times$ faster than the most accurate baseline. These results highlight the suitability of INR-based models for accurate and scalable analysis of myocardial strain in large CMR datasets. \url{www.github.com/andrewjackbell/Displacement-INR}

\end{abstract}

\section{Introduction}

The assessment of cardiac function is vital for diagnosing and managing cardiovascular disease. Myocardial strain, the relative deformation of the heart muscle, provides an early indication of left ventricular (LV) dysfunction~\cite{smisethMyocardialStrainImaging2016,hallidayAssessingLeftVentricular2021,xuStateoftheartMyocardialStrain2022} and independently predicts adverse outcomes~\cite{chadalavadaMyocardialStrainMeasured2024,xueAutomatedInLineArtificial2022,mordiCombinedIncrementalPrognostic2015}. Cardiac magnetic resonance (CMR) feature tracking is widely used for strain analysis, but overlooks intramyocardial motion by considering only tissue boundaries, making regional strain measures unreliable~\cite{buciusComparisonFeatureTracking2020}. 

CMR tagging imposes material `grid' features on the myocardium that can be tracked over the cardiac cycle. Tagging has been extensively validated~\cite{youngValidationTaggingMR1993,limaAccurateSystolicWall1993}, but requires time-consuming manual analysis, motivating the use of automated methods to rapidly and accurately quantify motion and strain for large imaging cohorts. Deep learning methods have been applied to tag tracking~\cite{ferdianFullyAutomatedMyocardial2020,yeDeepTagUnsupervisedDeep2021,loecherUsingSyntheticData2021} but are limited by low output resolution, image artefacts (e.g. tag fading) or slow inference.

Implicit neural representations (INRs) have been used in medical imaging for a variety of tasks to implicitly learn continuous functions from discrete training examples. Applications include super resolution~\cite{wuArbitraryScaleSuperResolution2023}, shape reconstruction~\cite{amiranashviliLearningShapeReconstruction2022}, registration~\cite{tianNePhiNeuralDeformation2024,wolterinkImplicitNeuralRepresentations2022,zimmerGeneralisedNeuralImplicit2024} and motion tracking~\cite{alvarez-florezDeepLearningAutomatic2024,arratialopezWarpPINNCineMRImage2023}. We applied INRs to represent a spatio-temporal displacement function of short-axis CMR tagging, and conditioned the network on latent codes of image frames. 

\subsection*{Contributions}

\begin{enumerate}
    \setlength{\itemsep}{8pt}
    \item We propose a novel INR method for motion tracking in CMR Tagging.

    \item The INR learns a realistic, continuous (arbitrarily resolvable) displacement function from sparse tracking data and myocardial incompressibility loss. 

    \item Our method achieves the lowest tracking error and combined strain error on the UK Biobank dataset compared to three deep learning baselines

\end{enumerate}

\section{Related Work}

\subsection{Tag tracking in CMR}

Classical tag tracking methods use frequency-domain~\cite{osmanCardiacMotionTracking1999} or image-intensity (registration) approaches~\cite{ledesma-carbayoUnsupervisedEstimationMyocardial2008}, but require manual initialisation and are not robust to heterogeneous data. Deep learning approaches have been used to learn directly from image data: Ferdian et al. trained a convolutional neural network (CNN) to predict landmark points using short-axis tagging images~\cite{ferdianFullyAutomatedMyocardial2020}. This method is inherently constrained to track a sparse set of points, whereas Ye et al. trained a network based on VoxelMorph to learn deformation at pixel-resolution~\cite{yeDeepTagUnsupervisedDeep2021,balakrishnanVoxelMorphLearningFramework2019}. However, it assumes preserved image intensity between image frames and uses global smoothing to regularise, which may lead to underestimation of real cardiac deformation. Loecher et al. trained a 3D CNN to predict motion paths from synthetic tagging patches over time~\cite{loecherUsingSyntheticData2021}. However, predicting displacements from patches is computationally expensive and may lead to irregular motion. 

Implicit Neural Representations have the potential to overcome these limitations by learning displacements from tracked points implicitly, while enforcing physiological constraints such as incompressibility via automatic differentiation. 

\subsection{Implicit Neural Representations}

In motion tracking, given two images $I_0$ and $I_t$ we aim to learn a function $f$ mapping material coordinate $X\in\mathbb{R}^d$ in the reference image $I_0$ to a displacement vector $u$:
\begin{equation}\label{eq:baseinr}
    u = f(X) \in \mathbb{R}^d,
\end{equation}
such that $X'=X+u$ is the new location of the material point in $I_t$. With implicit neural representations, we can optimise a multi-layer perceptron (MLP) to fit $f$ using
discrete training examples $(X,u)$ (supervised) or by warping $I_0$ by $u$ and maximising  $NCC(I_t, I_0 \circ u)$, where $NCC$ is the normalised cross-correlation (self-supervised).

In cardiac motion tracking, previous works have used INRs of the form given by equation \ref{eq:baseinr}, in the self-supervised (registration) setting~\cite{arratialopezWarpPINNCineMRImage2023,alvarez-florezDeepLearningAutomatic2024}. However, they share two limitations: (1) Network optimisation happens at inference-time (2) As in~\cite{yeDeepTagUnsupervisedDeep2021}, image similarity loss assumes preservation of image intensity between tracked frames.

The first limitation can be addressed by generalising the INR to a population, given some conditioning factor $Z\in\mathbb{R}^\ell$ unique to each case, such that

\begin{equation} \label{eq:general_inr}
    u=f(X, Z)\in \mathbb{R}^d,
\end{equation} 
where $Z$ is vector of size $\ell$. This approach has been used in other registration tasks~\cite{tianNePhiNeuralDeformation2024,zimmerGeneralisedNeuralImplicit2024,lowesImplicitNeuralRepresentations2025}.

The second limitation is particularly relevant in CMR tagging, where tag intensities fade over the cardiac cycle. To address these limitations, we train a generalised INR with tracked points as supervision, and learn the image latents end-to-end. 

\section{Method}

\begin{figure}
    \centering
    \includegraphics[width=\linewidth]{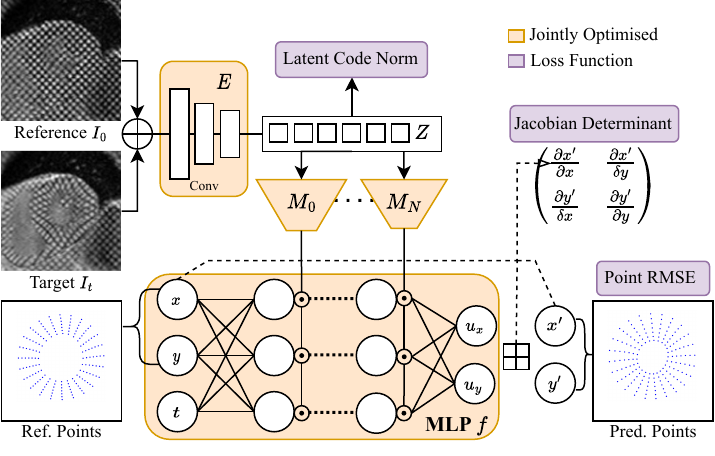}
    \caption{Method Overview. Convolutional encoder $E$ takes images $I_0$ and $I_1$ giving latent code $Z$, which is used to condition an MLP $f$ (via modulation networks $M_i$) that learns displacement between $I_0$ and $I_t$. $E$, $f$ and $M_i$ are jointly optimised using three losses shown.}
    \label{fig:architecture}
\end{figure}

 Our INR consists of an MLP $f$ that learns a continuous displacement function conditioned on image-specific latent codes. Given reference coordinate $X$ in $I_0$, latent code $Z$ and time coordinate $t$, $f$ predicts displacement vector $u$, giving the deformed coordinate $X'=X+u$ in $I_t$ (eq.~\ref{eq:general_inr}). The MLP $f$ uses sine activations~\cite{sitzmannImplicitNeuralRepresentations2020} with layer-wise modulation~\cite{mehtaModulatedPeriodicActivations2021} to condition the network on $Z$. For hidden layer $i$ with dimension $L$, the features $h_i\in\mathbb{R}^{L}$ are given by:

\begin{equation*}
    h_i = a_i \odot \sin( \omega (W_i h_{i-1} + b_i)),
\end{equation*}
where $W_i\in \mathbb{R}^{L \times L}$ is the weight matrix, and $b_i \in \mathbb{R}^L$ are the biases of $f$, $\omega \in\mathbb{R}$ is the frequency hyperparameter and $a_i\in\mathbb{R}^L$ are the modulation weights derived from $Z$.

To extract latent code $Z$ from image pair ($I_0, I_t$) we used a CNN encoder $E$: $Z=E(I_0,I_t)\in \mathbb{R}^\ell$ to learn a lower dimensional image representation. To adaptively condition $f$ on $Z$, modulation weights $a_i$ for each layer are given by a layer-specific modulation network $M_i$: $a_i=M_i(Z)\in\mathbb{R}^L$, where each $M_i$ is an MLP with a single hidden layer and ReLU activation.

\subsection{Training Objective}

To train this architecture, we optimised all components ($f$,$E$,$M_i$) end-to-end using a combination of three losses: The supervised position loss $\mathcal{L}_{pos}$ is the mean squared error between tracked points $X_i=(x_i,y_i)$ and predicted points $X'_i=(x_i',y_i')$ at target frame $I_t$:

\begin{equation*}
    \mathcal{L}_{pos}=\frac{1}{N}\sum_{i=1}^{N}\left[(x'_i-x_i)^2+(y'_i-y_i)^2 \right].
\end{equation*}

Latent code loss $\mathcal{L}_Z$ penalizes large magnitudes in $Z$:
\begin{equation*}
\mathcal{L}_Z = \frac{1}{\ell} \sum_{i=1}^{\ell} z_i^2.
\end{equation*}

The Jacobian loss $\mathcal{L}_J$ penalizes changes in local volume to enforce the approximate incompressibility of the myocardium~\cite{arratialopezWarpPINNCineMRImage2023}, providing a more suitable regularisation than global smoothing:
\begin{equation*}
\mathcal{L}_J = \frac{1}{N}\sum_{i=1}^N \left|1 - \det(J_i)\right|,\ \
J_i = \left. \frac{\partial X'}{\partial X}\right|_{X_i}
\end{equation*}

The weighted total loss is given by $
    \mathcal{L}= \mathcal{L}_{pos}+\alpha\mathcal{L}_J +\beta\mathcal{L}_Z,$
where $\alpha$ and $\beta$ are regularisation weights. 

\section{Experiments}

\subsection{Data}

The dataset used in our experiments consisted of 4508 cases in the UK Biobank (UKBB), each with three short-axis slices, randomly split into 3244, 812, and 452 cases for training, validation and testing, respectively. Details of acquisition and manual analysis were described in~\cite{ferdianFullyAutomatedMyocardial2020}. Observers manually adjusted a finite-element model consisting of bicubic Bézier curves to the LV myocardium at ED, using the CIM WARP software~\cite{youngTrackingFiniteElement1995}. Non-rigid registration deformations were manually corrected at ES and the final frame. A consistent set of 168 landmark positions was uniformly sampled from the model at each frame, giving tracked points which were used as ground truth for training and evaluation. 40 cases were analysed by two observers to calculate inter-observer statistics.

\subsection{Baseline Implementations}

\subsubsection{BioTag}~\cite{ferdianFullyAutomatedMyocardial2020} was previously trained on the same UK Biobank training set, with tracked points as supervision. We evaluated this method on the test set using the pre-trained model. Importantly, this method predicts points at ED and thus incurs an additional tracking error on this frame. However, strain estimates are directly comparable to other methods.

\subsubsection{SynthTag}~\cite{loecherUsingSyntheticData2021} was previously trained on a large synthetic patch dataset. We extracted patches surrounding each reference point and used the pre-trained model to predict a motion path for each point, yielding predicted points for all other frames. 

\subsubsection{DeepTag}~\cite{yeDeepTagUnsupervisedDeep2021} was retrained (unsupervised) on the UK Biobank training cases with parameters left unchanged from the paper. At evaluation time, we cropped images using BioTag's ROI localization network and applied the retrained model to predict deformation fields. The reference points at end-diastole were warped by the deformation fields at each frame to yield predicted points.
 
\subsection{INR Implementation}

The MLP $f$ consisted of 3 hidden layers, each with 256 nodes, sine activations, and layer-wise modulations. The encoder $E$ was a 2D CNN with 5 layers and progressively increasing filter sizes, each using a $3\times3$ kernel and stride 2. The encoder takes a 2-channel ($I_0\oplus I_t$) image of size $128\times128$ and the gives latent code $Z$ of size 32. Three modulation networks $M_i$, each consisting of an MLP with one hidden layer of size 256 and ReLU activation, take $Z$ and give modulation weights $a_i$ of size 256.

We trained our INR method on the UK Biobank with cropped short-axis tagging images and tracked points as supervision. All coordinates were normalised to the range $[-1,1]$ and intensity values in $[0,1]$. At evaluation time, BioTag's ROI localisation network was used to get cropped frames. Hyperparameters were chosen based on small-scale experiments on the validation set: $\alpha=1e^{-3}$, $\beta=1e^{-4}$, $\omega=15$, batch size was 4 and learning rate was $1e^{-4}$ (Adam Optimiser). Training was performed using Pytorch on an Nvidia RTX 4090 GPU for 14 epochs, lasting approximately $100$ minutes.

\subsection{Evaluation}

Point tracking error was calculated as the root mean squared Euclidean distance between the corresponding predicted and tracked points. 

Global Lagrangian strains were calculated as the inter-frame differences of neighbouring point distances. The local end-systolic strain $\epsilon$ for a pair of neighbouring points is given by:
\begin{equation}
    \epsilon= \frac{L_{ES}-L_{ED}}{L_{ED}},
\end{equation}
where $L_{ES}$ and $L_{ED}$ are the distances between the two points at end-systole and end-diastole respectively.

Global circumferential strain (GCS) was defined as the mean of local strains between neighbouring points around the circumference. Global radial strain (GRS) was similarly estimated from points along the radial direction. Strains were compared to manual ground truth by calculating biases (mean signed differences) and errors (mean absolute differences) between predicted and tracked reference strains.

\section{Results}

\begin{figure}
    \centering
    \begin{tabular}{ccc}
    End-Systolic Frame & Tracked Points & BioTag\\
    
    \includegraphics[width=0.32\linewidth]{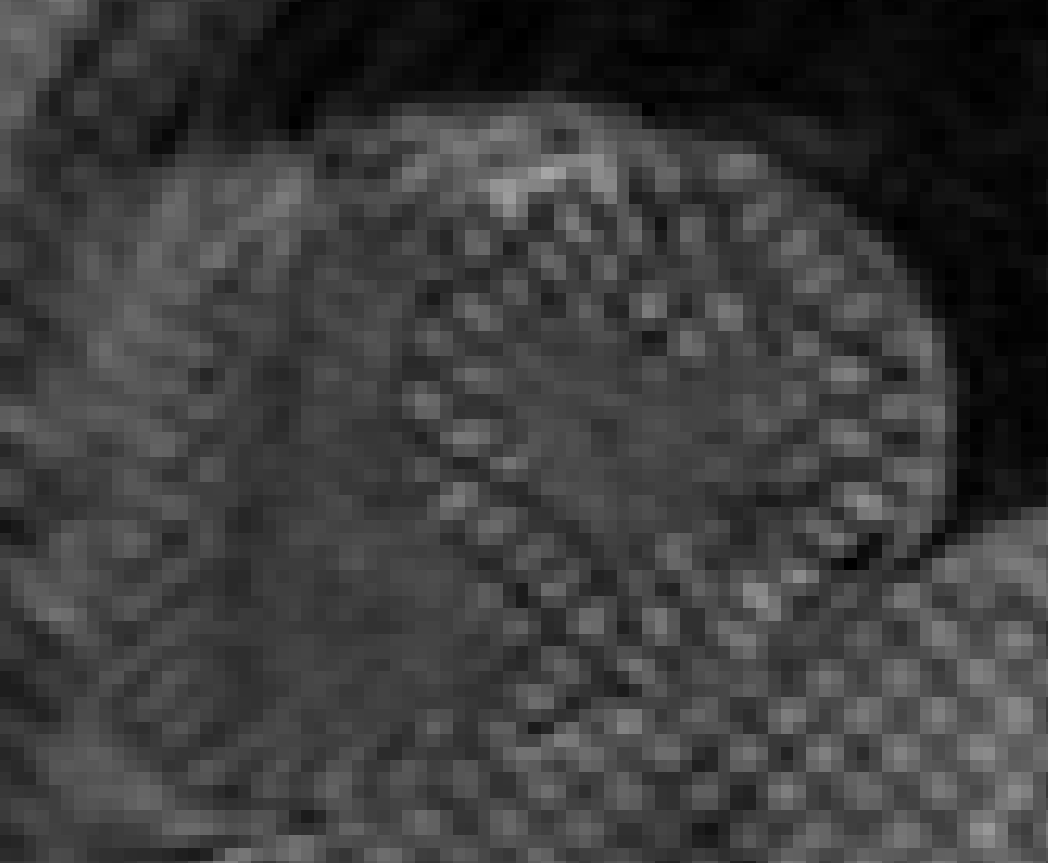} &
    \includegraphics[width=0.32\linewidth]{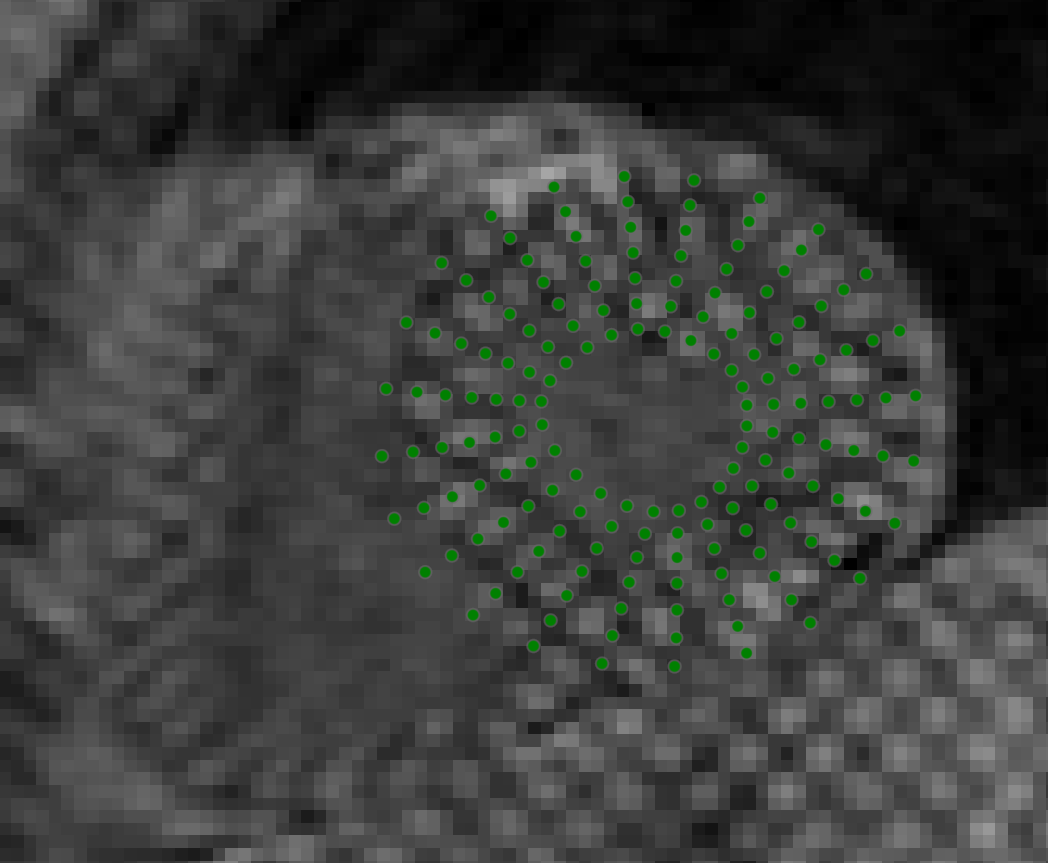} &
    \includegraphics[width=0.32\linewidth]{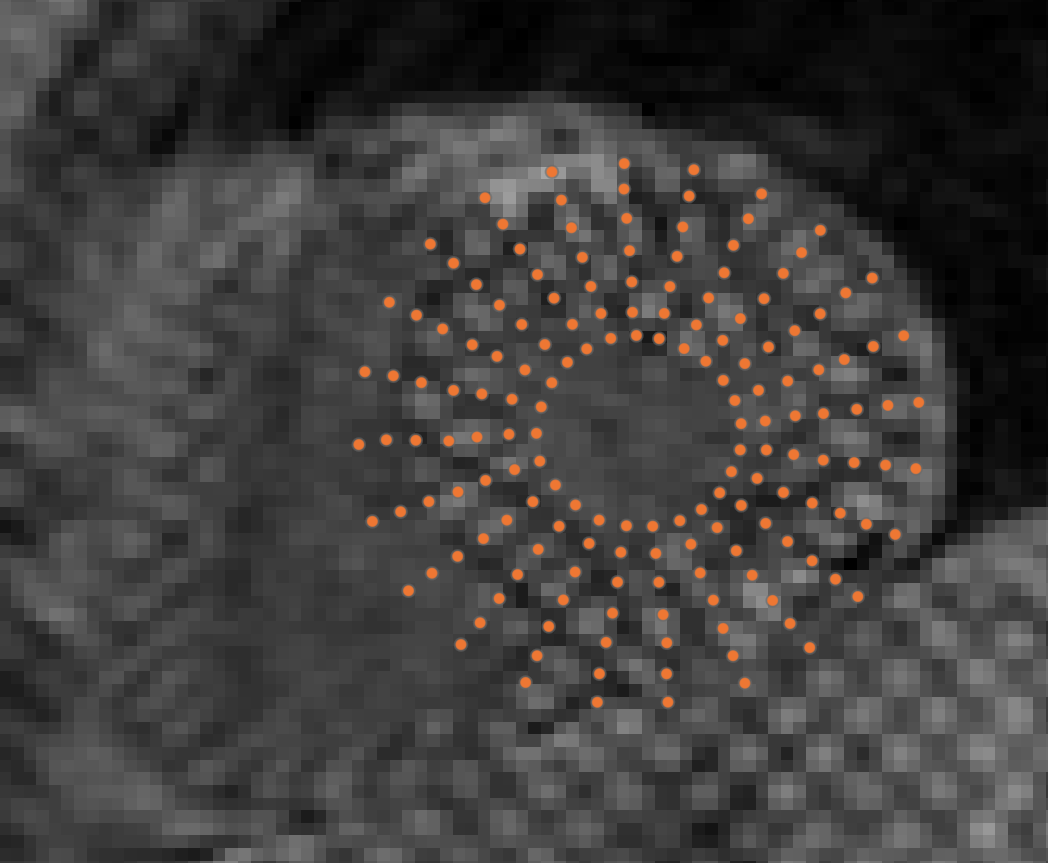} 
    \\
    DeepTag & SynthTag & INR (Ours)\\
    \includegraphics[width=0.32\linewidth]{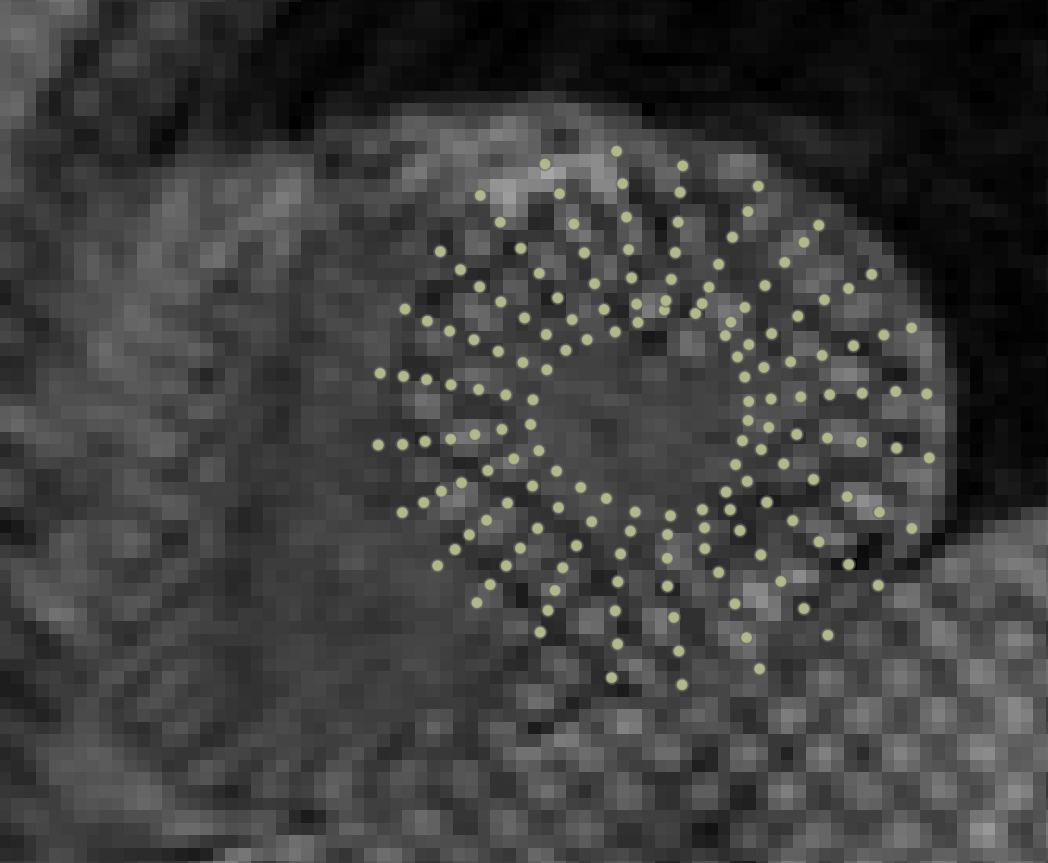}&
    \includegraphics[width=0.32\linewidth]{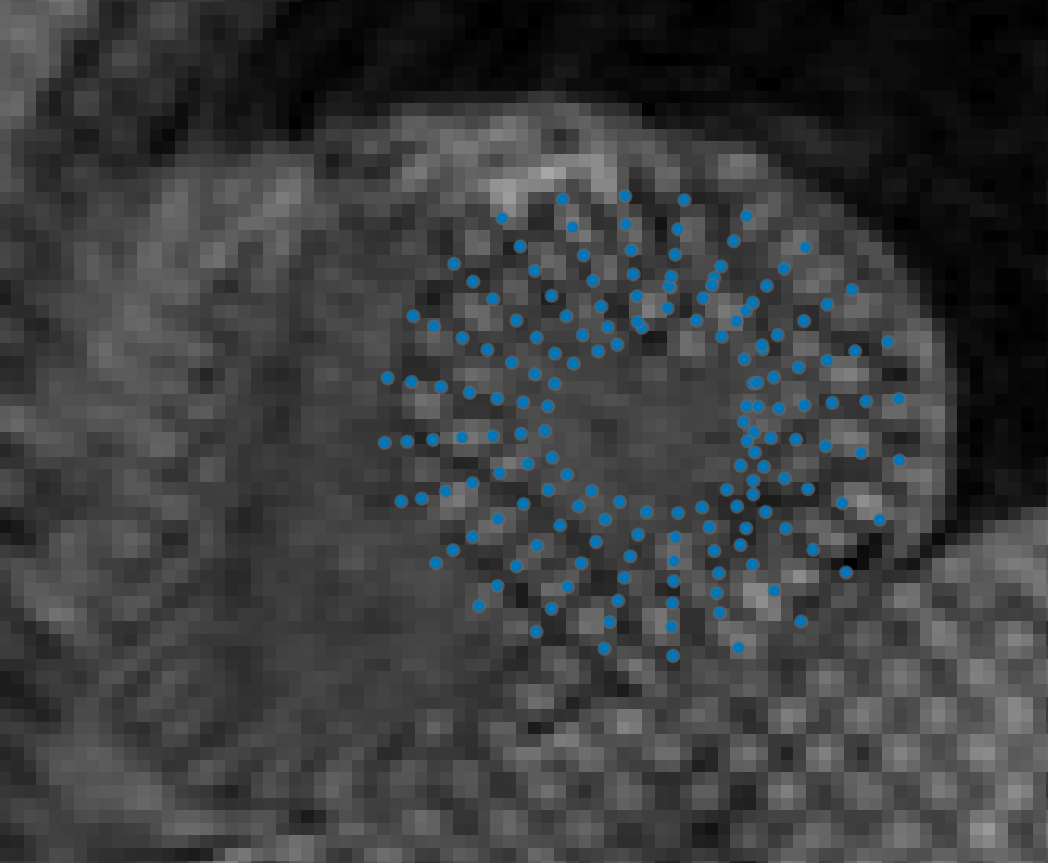} &
    \includegraphics[width=0.32\linewidth]{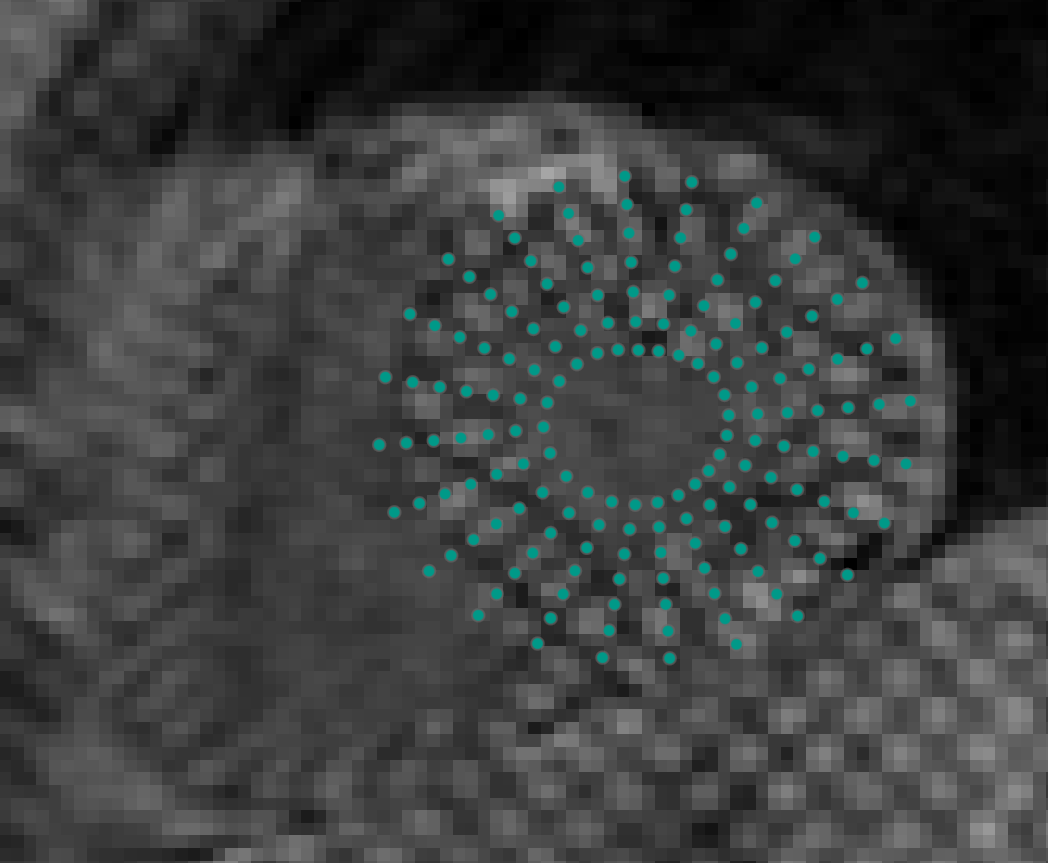}
        
    \end{tabular}
    \caption{Mid-ventricular end-systolic frame from the test dataset, with predicted points overlayed for each method}
    \label{fig:example_tracking}
\end{figure}

\begin{table}
    \setlength{\tabcolsep}{6pt}
    \centering
    \caption{Summary of performance between automatic and manual methods}
    \begin{tabular}{llccccc} 
    \toprule
    \textbf{}&
    \textbf{Method} & \makecell{\textbf{Point Error} \\ \textbf{(mm)}$\downarrow$} & \makecell{\textbf{Strain}\\\textbf{(\%)}} & \makecell{\textbf{Bias} \\ \textbf{(\%)}} & \makecell{\textbf{Error}\\ \textbf{(\%)}$\downarrow$} & \makecell{\textbf{Speed} \\\textbf{(slices/s)$\uparrow$}}  \\ 
    \midrule
        GCS & \textit{Manual* }& \textit{3.59} & \textit{-22.2} & \textit{1.83 ± 2.15} & \textit{2.38} & \textit{0.002} \\ 
        \textbf{} & BioTag & 4.26  & -21.8 & \textbf{0.42} ± 3.74 & 2.83 & 100 \\ 
        \textbf{} & DeepTag & 2.69 & -15.2 & 7.01 ± 3.66 & 7.06 & 250\\ 
        \textbf{} & SynthTag & 2.17 &-23.5 & -1.23 ± \textbf{3.04} & \textbf{2.54} & 3.29\\ 
        \textbf{} & INR (Ours) & \textbf{2.14} & -21.4 & 0.82 ± 3.70 & 2.86 & \textbf{1,250}\\ 
        \midrule
        GRS & \textit{Manual*} &\textit{3.59}& \textit{19.3} & \textit{1.60 ± 4.62} & \textit{3.80} & \textit{0.002} \\ 
        \textbf{} & BioTag & 4.26 & 18.7 & \textbf{-0.59} ± 8.51 & 6.64 & 100 \\ 
        \textbf{} & DeepTag &2.69 & 17.7 & -1.53 ± 7.33 & \textbf{5.85}& 250\\ 
        \textbf{} & SynthTag &2.17 & 12.7 & -6.58 ± \textbf{7.01} & 7.81& 3.29 \\ 
        \textbf{} & INR (Ours)& \textbf{2.14} & 16.4 & -2.81 ± 7.64 & 6.42 & \textbf{1,250} \\ 
        \bottomrule
    \end{tabular}
    *Manual statistics based on inter-observer differences on 40 cases
    
    \label{table:strain_errors}
\end{table}

\begin{table}[b]
    \centering
    \setlength{\tabcolsep}{6pt}
    \caption{Ablation results on Jacobian weight $\alpha$}
    
\begin{tabular}{lccccc}
    \toprule
    \textbf{$\alpha$} & \makecell{\textbf{Point Error}\\(mm)$\downarrow$}   &\makecell{ \textbf{GCS Bias} \\ (\%)} & \makecell{\textbf{GCS Error}\\(\%)$\downarrow$}   & \makecell{ \textbf{GRS Bias} \\ (\%)} & \makecell{\textbf{GRS Error}\\(\%)$\downarrow$}  \\ 
    \midrule
    $0$      & \textbf{2.12} & \textbf{0.64} & \textbf{2.84} & -4.38          & 7.03       \\ 
    $0.001$  & 2.14          & 0.82          & 2.86          & -2.81          & \textbf{6.42}  \\ 
    $0.005$  & 2.20          & 1.22          & 3.02          & \textbf{-0.40} & 6.66     \\ 
    $0.01$   & 2.27          & 0.86          & 3.06          & 3.05           & 7.75  \\ 
    $0.1$    & 2.59          & 3.17          & 4.31          & 6.92           & 10.47       \\
    \bottomrule
    
\end{tabular}

    \label{table:ablation}
\end{table}

\section{Discussion}

Motion tracking in CMR tagging is a challenging task due to motion blur, tag fading and through-plane motion. We introduced a lightweight, generalised INR method to predict physiologically plausible displacement in short-axis tagging images at arbitrary resolution. Table~\ref{table:strain_errors} shows that our method had the lowest tracking error (2.14mm), followed by SynthTag (2.17mm), and DeepTag (2.69mm). Fig.~\ref{fig:example_tracking} shows an example case with tracking differences between methods. Our method appears to agree with manual tracking, but shows a more constricted endocardium. The Jacobian loss promotes inward tracking of endocardial points to preserve volume --- a behavior consistent with physiological myocardial motion. Despite having comparable mean tracking error, SynthTag tracking appears irregular in some cases, with overlapping points. This is because point predictions are made independently based on image patches, which are sensitive to local artefacts or noise. DeepTag tracking was smoother due to regularisation, but had greater mean tracking errors. BioTag incurs larger errors as it also predicts points at the reference frame. In this example, the outermost points lie too far outside of the epicardial boundary. In terms of tracking performance and regularity, our INR method demonstrates improvements over baseline methods.

As shown in table \ref{table:strain_errors}, we found that SynthTag had the lowest GCS error (2.54\%), but also underestimated GRS (-6.58\% bias) and had the largest GRS error (7.81\%). DeepTag had the best GRS error (5.85\%), but had the largest GCS error. BioTag had excellent biases (0.42\%, -0.59\%), but had the largest variances (3.74\%, 8.51\%) for GCS and GCS respectively. Our INR method is the most balanced approach with the lowest combined error (2.86\%, 6.42\%) and relatively small biases (0.82\%, -2.81\%) for GCS and GRS respectively. Table~\ref{table:ablation} shows the effect of Jacobian weight $\alpha$ on tracking performance and strain differences. We found that using no regularisation marginally improved tracking accuracy and GCS Error at the expense of increased GRS error. Using $\alpha>0.001$ reduced performance on all metrics. Therefore we believe using $\alpha=0.001$ is optimal for this task as GCS errors are marginally increased while substantially reducing GRS errors.

Our results indicate that the INR method is well suited to this task, being able to predict realistic and accurate deformation, while overcoming limitations of previous works. However, two main limitations of the study should be considered: Firstly, the manually tracked points are subject to inter-observer differences that reflect a level of tracking imprecision. Secondly, our evaluations were performed on a single dataset containing mostly healthy volunteers. In the future, synthetic data will be explored to mitigate the need for manual tracking and additional loss terms will be used which exploit INR differentiability. Finally, the method will be validated on clinical datasets with significant cardiovascular disease to ensure accuracy in the presence of LV dysfunction. 

\section{Conclusion}

We introduced a generalised implicit neural representation (INR) for efficient and accurate tracking of intramyocardial motion from CMR tagging. Our method learns a continuous displacement field without inference-time optimisation, using sparse supervision and a loss function that enforces myocardial incompressibility. Experiments on the UK Biobank showed that our method outperformed existing deep learning baselines in both tracking accuracy and strain analysis. These results demonstrate the potential of INR-based models for scalable and accurate myocardial motion analysis in large cohorts.

\newpage

\bibliographystyle{splncs04}
\bibliography{myocardial_inr}

\end{document}